\documentclass[10pt,twocolumn,letterpaper]{article}

\usepackage[pagenumbers]{cvpr} 

\usepackage{graphicx}
\usepackage{amsmath}
\usepackage{amssymb}
\usepackage{booktabs}

\usepackage{multirow}
\usepackage[normalem]{ulem}
\useunder{\uline}{\ul}{}

\usepackage{amsmath}
\usepackage{amssymb}
\usepackage{bm}

\usepackage[pagebackref,breaklinks,colorlinks]{hyperref}

\usepackage[capitalize]{cleveref}
\crefname{section}{Sec.}{Secs.}
\Crefname{section}{Section}{Sections}
\Crefname{table}{Table}{Tables}
\crefname{table}{Tab.}{Tabs.}

\begin{document}

\title{Event-based Monocular Dense Depth Estimation with Recurrent Transformers}

\author{Xu Liu
\and
Jianing Li
\and
Xiaopeng Fan
\and
Yonghong Tian
}
\maketitle

\begin{abstract}
Event cameras, offering high temporal resolutions and high dynamic ranges, have brought a new perspective to address common challenges (e.g., motion blur and low light) in monocular depth estimation. However, how to effectively exploit the sparse spatial information and rich temporal cues from asynchronous events remains a challenging endeavor. To this end, we propose a novel event-based monocular depth estimator with recurrent transformers, namely EReFormer, which is the first pure transformer with a recursive mechanism to process continuous event streams. Technically, for spatial modeling, a novel transformer-based encoder-decoder with a spatial transformer fusion module is presented, having better global context information modeling capabilities than CNN-based methods. For temporal modeling, we design a gate recurrent vision transformer unit that introduces a recursive mechanism into transformers, improving temporal modeling capabilities while alleviating the expensive GPU memory cost. The experimental results show that our EReFormer outperforms state-of-the-art methods by a margin on both synthetic and real-world datasets. We hope that our work will attract further research to develop stunning transformers in the event-based vision community. Our open-source code can be found in the supplemental material.

\end{abstract}

\section{Introduction}
\label{sec:intro}
Monocular depth estimation~\cite{dong2022towards, laga2020survey}, is one of the critical and challenging topics, which support widespread vision applications in a low-cost effective manner. In fact, conventional frame-based cameras have presented some shortcomings for depth estimation in challenging conditions (e.g., motion blur and low light)~\cite{hu2014joint, wang2021regularizing}. Recently, event cameras~\cite{gallego2022event, posch2014retinomorphic}, offering high temporal resolutions and high dynamic ranges, having been attempted to address these common challenges~\cite{gehrig2021combining, hidalgo2020learning, zhu2019unsupervised, chaney2019learning, haessig2019spiking, wang2021dual, gallego2018unifying, rebecq2018emvs, gallego2019focus, cui2022dense, cho2021eomvs, kim2016real}. However, a key question remains: \emph{How to effectively exploit the sparse spatial information and rich temporal cues from asynchronous events to generate dense depth maps?}

\begin{figure}[t]
	\begin{subfigure}[b]{\linewidth}
		\centering
		\includegraphics[width=8.30cm]{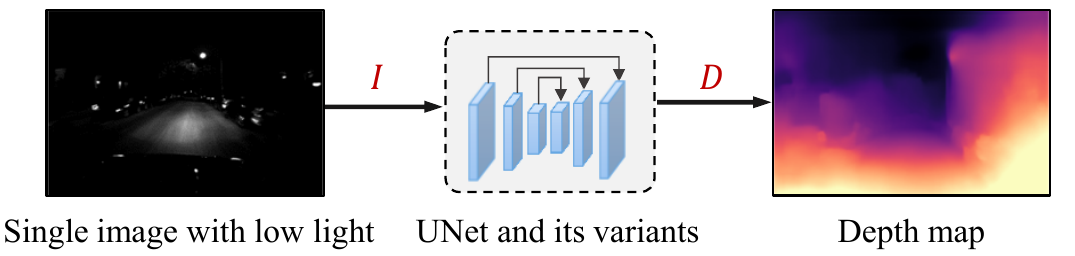}
		\caption{Feed-forward frame-based monocular depth estimators~\cite{li2018megadepth, wang2021regularizing, ranftl2021vision}}
	\end{subfigure}
	
	\begin{subfigure}[b]{\linewidth}
		\centering
		\includegraphics[width=8.32cm]{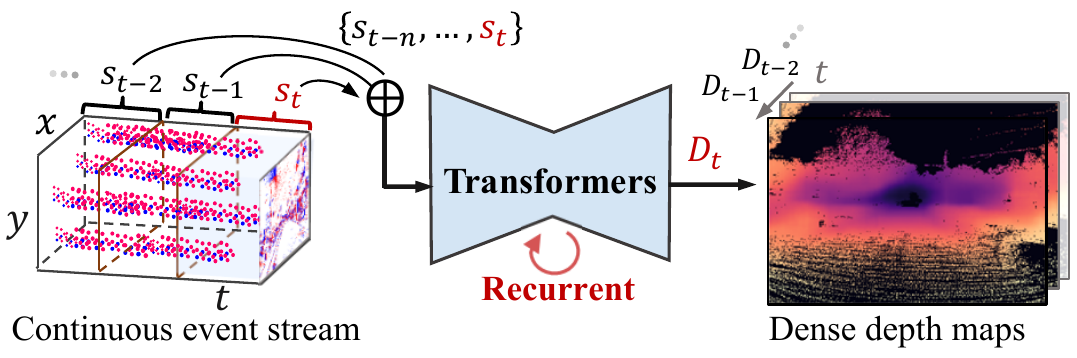}
		\caption{Event-based monocular depth estimation with our EReFormer}
	\end{subfigure}
	\vspace{-0.30cm}
	\caption{Monocular depth estimation in challenging condition. (a) Feed-forward frame-based monocular depth estimators~\cite{li2018megadepth, wang2021regularizing, ranftl2021vision} fail to generate a high-quality depth map by processing each image with low light. (b) Our EReFormer is a pure transformer with a recursive mechanism, which can convert continuous event stream into high-quality dense depth maps via modeling global context information and leveraging rich temporal cues.}
	\label{fig:intro}
	\vspace{-0.36cm}
\end{figure}

For \emph{\textbf{spatial modeling}}, the mainstream event-based monocular depth estimators~\cite{gehrig2021combining, hidalgo2020learning, zhu2019unsupervised, wang2021dual} adopt CNN-based architectures, resulting in \emph{insufficiently exploiting global context information} from asynchronous sparse events. For instance, Zhu~\emph{et al.}~\cite{zhu2019unsupervised} design an unsupervised encoder-decoder network for semi-dense depth estimation. Further, the following works~\cite{gehrig2021combining, hidalgo2020learning, wang2021dual} present supervised training frameworks to generate dense depth maps based on UNet~\cite{ronneberger2015u}. Although these CNN-based learning methods achieve better performance than the model-based optimized approaches~\cite{gallego2018unifying, rebecq2018emvs, gallego2019focus, kim2016real}, they are not capable of utilizing the global spatial information due to the essential locality of convolution operations. More recently, Transformers~\cite{vaswani2017attention, ranftl2021vision, guizilini2022multi, junayed2022himode, wang2022learning} demonstrate appealing potential in modeling global context information for frame-based monocular depth estimation tasks. Yet, so far, there is still no transformer-based depth estimator for event cameras.

For \emph{\textbf{temporal modeling}}, most existing event-based monocular depth estimators run feed-forward models~\cite{zhu2019unsupervised, wang2021dual, chaney2019learning} or introduce RNN-based architectures~\cite{gehrig2021combining, hidalgo2020learning}, \emph{limiting their abilities to leverage temporal dependency}. More specifically, the feed-forward models~\cite{zhu2019unsupervised, wang2021dual, chaney2019learning} generate each depth map via independently processing a voxel grid, thus they have not yet utilized temporal cues from continuous event streams. Consequently, the lightweight recurrent convolutional architectures (e.g., ConvLSTM~\cite{hidalgo2020learning} and ConvGRU~\cite{gehrig2021combining}) are attempted to incorporate into UNet~\cite{ronneberger2015u} for modeling long-range temporal dependencies. However, the lack of spatio-temporal information interactions in CNN-based backbones may incur the bottlenecks of performance improvements. By contrast, Transformers effectively establish the interaction between spatial and temporal domains via the self-attention mechanism, they have demonstrated impressive performance in temporal sequence tasks~\cite{cao2022vdtr,li2021video}. Of course, such temporal Transformers require a large GPU memory cost, and their input within the batch also limits information in temporal features. Therefore, how to design a transformer-based monocular depth estimator that aims at leveraging rich temporal cues meanwhile alleviating the expensive GPU memory cost still remains open.

To address the aforementioned problems, this paper proposes a novel \textbf{e}vent-based monocular dense depth estimator with \textbf{re}current trans\textbf{former}s, namely \emph{\textbf{EReFormer}}, which is the first pure transformer-based encoder-decoder architecture with a recursive mechanism to process continuous event streams, as shown in Fig.~\ref{fig:intro}. In fact, the goal of this work is not to optimize transformer-based monocular depth estimators (e.g., DPT~\cite{ranftl2021vision}) on each event image. On the contrary, we aim at overcoming the following challenges: (i) \emph{\textbf{Global sparse spatial modeling}} - \emph{How do we design a transformer-based monocular depth estimator that effectively exploiting global context information from sparse events?} (ii) \emph{\textbf{Efficient temporal utilization}} - \emph{What is the transformer-based architecture that efficiently leverages rich temporal cues from continuous event streams?}

Toward this end, our EReFormer is designed to model global context information and long-range temporal dependencies from asynchronous events. More specifically, we first design a novel encoder-decoder transformer-based backbone using Swin Transformer blocks~\cite{liu2021swin} for event-based monocular depth estimation, which has better global context information modeling capabilities than CNN-based methods. Then, a \textbf{s}patial \textbf{t}ransformer \textbf{f}usion (STF) module is present as a skip connection to fuse multi-scale features in our EReFormer, which obtains richer spatial contextual information from sparse event streams. Finally, we propose a \textbf{g}ate \textbf{r}ecurrent \textbf{vi}sion \textbf{t}ransformer (GRViT) unit to leverage rich temporal cues from event streams. The core of GRViT is to introduce a recursive mechanism into transformers so that it can benefit the performance and alleviate the expensive GPU memory cost. The experimental results show that our EReFormer outperforms state-of-the-art methods by a large margin on both synthetic and real-world datasets (i.e., DENSE~\cite{hidalgo2020learning} and MVSEC~\cite{zhu2018multivehicle}). Our EReFormer also verifies that event cameras can perform robust monocular depth estimation even in cases where conventional cameras fail, e.g.,  fast-motion and low-light scenes.

In summary, the main contributions of this paper are summarized as follows:
\vspace{-\topsep}
\begin{itemize}
\setlength{\parskip}{0pt} \setlength{\itemsep}{0pt plus 1pt}
	\item We propose a novel \emph{pure transformer-based architecture} for event-based monocular depth estimation, which outperforms state-of-the-art methods in terms of depth map quality by a large margin.
	\item We establish a \emph{spatial transformer fusion module} to improve spatial global modeling capabilities via fusing multi-scale features from asynchronous sparse events. 
	\item We design a \emph{gate recurrent vision transformer unit} that incorporates a recursive mechanism into transformers, which can leverage rich temporal cues meanwhile alleviating the expensive GPU memory cost.
\end{itemize}
\vspace{-\topsep}

To the best of our knowledge, this is the first work to explore such a pure transformer to generate dense depth maps for a monocular event camera, which further unveils the versatility and transferability of transformers from conventional frames to continuous event streams.

\begin{figure*}[t]
  \centering
  \includegraphics[width=0.99\textwidth]{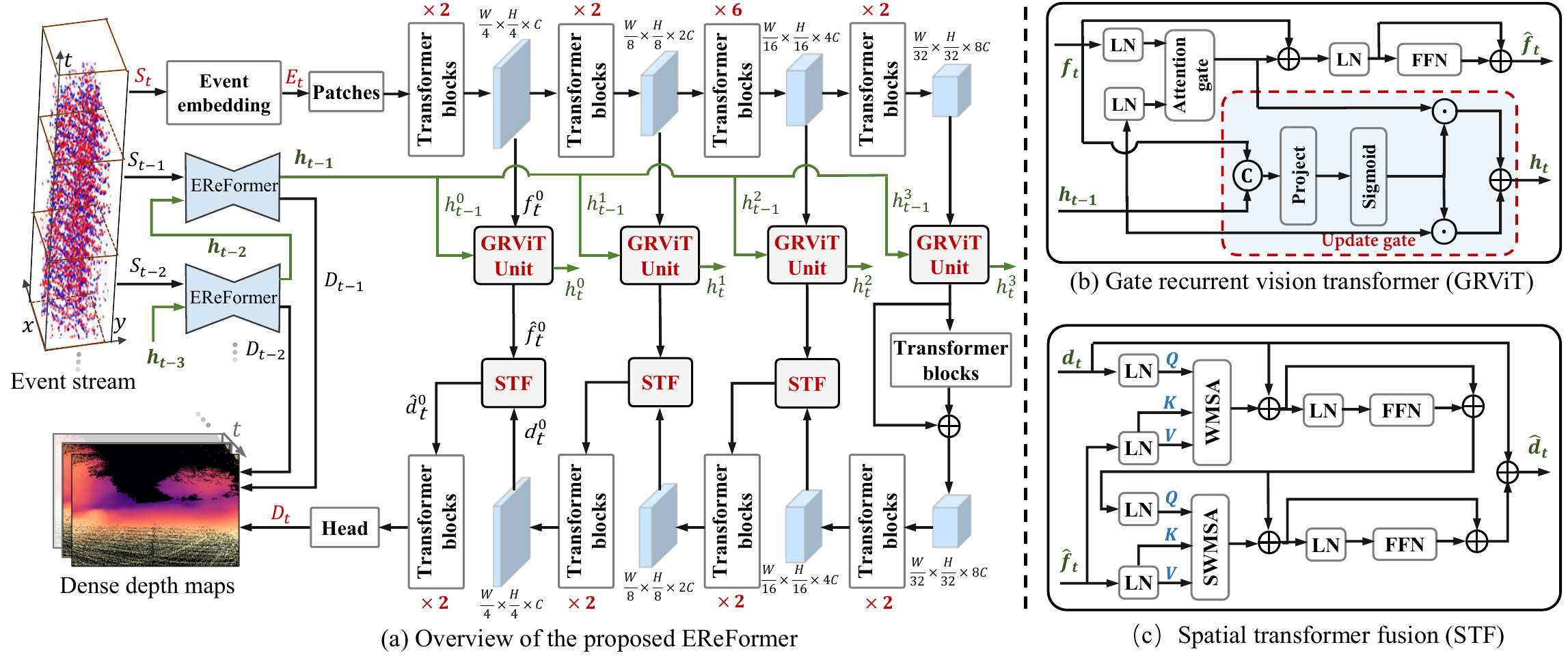}
  \caption{The structure of the proposed event-based monocular depth estimator with recurrent transformers. (a) The overall workflow of our EReFormer. The event stream is first converted into event embeddings~\cite{gehrig2019end} and then split into non-overlapping patches. Then, the patches are processed via an encoder-decoder sub-network with transformer blocks. (b) The proposed GRViT unit introduces a recursive mechanism into transformers to leverage temporal cues. (c) The designed STF module is a skip connection to fuse multi-scale features.}
  \label{fig:framework}
  \vspace{-0.30cm}
\end{figure*}

\section{Related Work}

\noindent \textbf{Event-based Monocular Depth Estimation.} Event cameras for monocular depth estimation has become increasingly popular in robot navigation~\cite{gallego2022event, falanga2020dynamic, mitrokhin2019learning}, especially involving low-latency obstacle avoidance and high-speed path planning. Early model-based works~\cite{gallego2018unifying, rebecq2018emvs, gallego2019focus, cui2022dense, cho2021eomvs, kim2016real} usually calculate both camera poses and depth maps via solving a non-linear optimization problem. Yet, these model-based optimized methods need to obtain camera poses or auxiliary sensor parameters (e.g., IMU). Recently, various learning-based methods~\cite{gehrig2021combining, hidalgo2020learning, zhu2019unsupervised, chaney2019learning, haessig2019spiking, wang2021dual} have been introduced to convert asynchronous events into depth maps. Although these CNN-based methods achieve promising results, some of these feed-forward models~\cite{ zhu2019unsupervised, chaney2019learning, wang2021dual} have not yet used rich temporal cues, and the lack of spatio-temporal information interactions in CNN-based backbones may limit performance improvements. Therefore, this work aims to effectively modeling spatio-temporal information from asynchronous events to generate dense depth maps.

\noindent \textbf{Monocular Depth Estimation with Transformers.} With the self-attention mechanism, transformer-based monocular depth estimators~\cite{ranftl2021vision, ji2021monoindoor, zhao2022monovit, guizilini2022multi, cheng2021swin, hwang2022self, han2022transdssl, agarwal2022depthformer} have achieved finer-grained and more globally coherent predictions than CNN-based methods. It is worth mentioning that DPT~\cite{ranftl2021vision} first leverages vision transformers instead of CNN-based backbones for dense depth prediction tasks. Subsequently, some studies~\cite{ji2021monoindoor, zhao2022monovit, hwang2022self, han2022transdssl} adopt transformers for self-supervised monocular depth estimation. Actually, these transformer-based architectures operate on each isolated image so that they do not directly process a continuous stream of asynchronous events. Inspired by the ability of transformers to model long-range temporal dependencies in video sequence tasks~\cite{cao2022vdtr,li2021video, yang2022recurring}, we design a gate recurrent vision transformer unit to leverage rich temporal cues.

\noindent \textbf{Transformers in Event-based Vision.} Some event-based vision tasks (e.g., event representation~\cite{sabater2022event}, video reconstruction~\cite{weng2021event}, denoising~\cite{alkendi2022neuromorphic}, object tracking~\cite{zhang2022spiking}, and object recognition~\cite{zhao2022transformer}) have sought to design transformer-based frameworks for better performance. For example, ET-Net~\cite{weng2021event} introduces transformers into CNN for event-based video reconstruction, which effectively models global context via the transformer-based module. Alkendi \emph{et al.}~\cite{alkendi2022neuromorphic} develop a hybrid GNN-Transformer model for event camera denoising. CTN~\cite{zhao2022transformer} presents a hybrid CNN-Transformer network for event-based data classification. However, these hybrid architectures are performance-oriented and cannot reveal the transferability of a pure transformer in monocular depth estimation. Thus, we design a pure transformer-based architecture for event-based monocular depth estimation.

\section{Problem Definition}
Event cameras, such as DVS~\cite{lichtsteiner2008128} and DAVIS~\cite{brandli2014240}, are bio-inspired vision sensors that respond to light changes with continuous event streams. Each event $\boldsymbol{e_{n}}$ can be described as a four-attribute tuple $(x_{n}, y_{n}, t_{n}, p_{n})$. Consequently, asynchronous events $\boldsymbol{S}$=$\left\lbrace  \boldsymbol{e_{n}} \right\rbrace _{n=1}^{N_{e}}$ are sparse and discrete points in the spatio-temporal window $\Gamma$. In general, a continuous event stream needs to be split into event temporal bins. Obviously, the temporal correlation lies in adjacent event temporal bins~\cite{li2022asynchronous}. However, the most existing event-based monocular depth estimators~\cite{zhu2019unsupervised, wang2021dual, chaney2019learning}, running a feed-forward frame-based model independently on each event image~\cite{maqueda2018event} or voxel grid~\cite{zhu2019unsupervised}, have not yet leveraged rich temporal cues. In this work, we focus on the knowledge gap and formulate this challenging issue called \emph{\textbf{event-based monocular depth estimation}} as follows.

Let $\left\lbrace S_{1}, ..., S_{T}\right\rbrace$ be event temporal bins separated from a continuous event stream $\boldsymbol{S}$, where $S_{t}\in \mathbb{R}^{W \times H \times \Delta t}$ is the $t$-th event temporal bin with the duration $\Delta t$. To make asynchronous events compatible with deep learning techniques~\cite{gehrig2019end}, event temporal bins need to convert into event embeddings $\boldsymbol{E}$=$\left\lbrace E_{1}, ..., E_{T}\right\rbrace$ by a kernel function $\mathcal{K}$, where $E_{t}\in \mathbb{R}^{W \times H \times C_{e}}$ is $t$-th event embedding with the channel number $C_{e}$. The goal of our monocular depth estimator is to learn a non-linear mapping function $\mathcal{M}$ to generate dense depth maps $\boldsymbol{D}$=$\left\lbrace D_{1}, ..., D_{T}\right\rbrace $ by exploiting the spatio-temporal information, it can be formulated as:
\begin{eqnarray}
\boldsymbol{D}= \mathcal{M}  ( \mathcal{K} (S_{1}),..., \mathcal{K}( S_{T} ) ),
\end{eqnarray}
where the proposed function $\mathcal{M}$ can leverage rich temporal cues from event temporal bins, and the parameter $T$ determines the length of utilizing temporal information.

Given the ground-truth depth maps $\bar{\boldsymbol{D}}$=$\left\lbrace \bar{D}_{1}, ..., \bar{D}_{T}\right\rbrace $, we minimize the loss function between the predicted depth map $D_{t}$ and the ground-truth $\bar{D}_{t}$ as follows:
\begin{eqnarray}
\hat{\mathcal{M}}=\mathop{\arg \min}_{\mathcal{M}} L_{\mathcal{M}} (\boldsymbol{D}, \bar{\boldsymbol{D}}) \triangleq \mathbb{E}_{t \in [1, T]}\left[d\left( D_t, \bar{D}_t\right)\right] ,
\end{eqnarray}
where $\mathbb{E}[\cdot]$ is an empirical expectation function, $d(\cdot,\cdot)$ is a distance metric, e.g., scale-invariant loss.

\section{Methodology}

\subsection{Network Overview}
This work aims at designing a novel event-based monocular depth estimator with recurrent transformers, termed \textbf{\emph{EReFormer}}, which can generate high-quality dense depth maps via modeling global context information and leveraging rich temporal cues. As shown in Fig.~\ref{fig:framework}\textcolor{red}{(a)}, our EReFormer mainly consists of three modules: \emph{transformer-based encoder-decoder}, \emph{spatial transformer fusion (STF) module}, and \emph{gate recurrent vision transformer (GRViT) unit}. More precisely, the event stream $\boldsymbol{S}$ is first split into event temporal bins $\left\lbrace S_{1}, ..., S_{T}\right\rbrace$, and each bin $S_{t}$ is converted into a 2D image-like representation $E_{t}$. we encode each bin into an event image~\cite{maqueda2018event} owning to an accuracy-speed trade-off. In fact, any event representation can be an alternative because our EReFormer provides a generic interface. Then, the transformer-based encoder, utilizing Swin transformer blocks~\cite{liu2021swin}, progressively extract multi-scale features via the downsampling operation. Meanwhile, the GRViT unit incorporates a recursive mechanism into transformers to model long-range temporal dependencies, which can leverage rich temporal cues meanwhile alleviating the expensive GPU memory cost. To further improve global spatial modeling capabilities, the STF module is designed as a skip connection to fuse multi-scale features. Finally, the corresponding decoder predicts fine-grained and globally coherent depth maps $\left\lbrace D_{1}, ..., D_{T}\right\rbrace$ using the hierarchical upsampling transformer blocks.

\subsection{Global Spatial Modeling with Transformers}

Due to the sparse and discrete attributes of asynchronous events, it is difficult to extract effective global spatial information from the local space using CNN-based models. To overcome this challenge, we design a pure transformer-based encoder-decoder and a STF module to model global spatial information from asynchronous events.

\textbf{Transformer Encoder.} To enhance the global information learning ability under different scale features, we design a hierarchical network as the backbone, which uses multiple Swin transformer blocks~\cite{liu2021swin} to implement spatial downsampling. Specifically, a 2D image-like representation $E_{t}\in \mathbb{R}^{W \times H \times C_{e}}$ is first split into non-overlapping patches with the size 4$\times$4 and then projected to tokens with the dimension $C$ by a patch embedding layer. Furthermore, all tokens are input to four transformer layers with different block numbers (i.e., 2, 2, 6, and 2), and each transformer layer 
performs the downsampling operation to reduce the spatial resolution and increases the channel number with a factor of 2 (see the top of Fig.~\ref{fig:framework}\textcolor{red}{(a)}). As a result, the output feature maps of four transformer layers from low level to high level are $f_{t}^{0} \in \mathbb{R}^{\frac{H}{4} \times \frac{W}{4} \times C}$, $f_{t}^{1} \in \mathbb{R}^{\frac{H}{8} \times \frac{W}{8} \times 2C}$, $f_{t}^{2} \in \mathbb{R}^{\frac{H}{16} \times \frac{W}{16} \times 4C}$, and $f_{t}^{3} \in \mathbb{R}^{\frac{H}{32} \times \frac{W}{32} \times 8C}$.

\textbf{Transformer Decoder.} As a symmetrical architecture, the corresponding decoder is also a hierarchical network with four transformer layers. Unlike the encoder, each layer of the decoder adopts two Swin transformer blocks (see the bottom of Fig.~\ref{fig:framework}\textcolor{red}{(a)}). For the first three layers, the output feature maps are $d_{t}^{2} \in \mathbb{R}^{\frac{H}{16} \times \frac{W}{16} \times 4C}$, $d_{t}^{1} \in \mathbb{R}^{\frac{H}{8} \times \frac{W}{8} \times 2C}$, and $d_{t}^{0} \in \mathbb{R}^{\frac{H}{4} \times \frac{W}{4} \times C}$, respectively. In detail, each layer first increases the channel number and then decreases the spatial resolution via the patch splitting operation. After that, the last transformer layer further refines the feature map $d_{t}^{0}$, and a task-specific head is implemented to predict a dense depth map $D_{t}$ by the sigmoid function.

\textbf{Spatial Transformer Fusion.} Most event-based monocular depth estimators~\cite{hidalgo2020learning,wang2021dual,zhu2019unsupervised} adopt the aggregation operation (e.g., ADD or CONCAT) as a skip connection to fuse multi-scale features. However, these fusion strategies insufficiently exploit global spatial context information from sparse asynchronous events. Thus, we propose a spatial transformer fusion (STF) module to overcome this limitation via cross-attention learning.

Our STF module mainly consists of two core transformer blocks, namely regular window-based and shifted window-based multi-head self-attention (i.e., WSMA and SWMSA~\cite{liu2021swin}). As illustrated in Fig.~\ref{fig:framework}\textcolor{red}{(c)}, WSMA and SWMSA perform the cross-attention operation with a residual connection, respectively. Take WMSA for instance, we use the decoded feature map $d_{t}$ to generate query ($Q_{t}$), and utilize the output $\hat{f}_{t}$ of GRViT (see Section~4.3) to generate the key ($K_{t}$) and value ($V_{t}$). Taking the triplet (i.e, $Q_{t}$, $K_{t}$, and $V_{t}$) as the input, our STF module progressively models spatial contextual information and outputs the cross-attention feature map $\bar{d}_{t}$. Finally, the fused feature map $\hat{d}_{t}$ is obtained by a residual connection to integrate $d_{t}$ and $\bar{d}_{t}$. Thus, our STF module can be formulated as follows:
\begin{equation}
\begin{array}{l}
\tilde{d}_{t}=\operatorname{\scriptstyle WMSA}\left(d_{t}, \hat{f}_{t}\right)+\operatorname{FFN}\left(\operatorname{\scriptstyle WMSA}\left(d_{t}, \hat{f}_{t}\right)\right)\\
\bar{d}_{t}=\operatorname{\scriptstyle SWMSA}\left(\tilde{d}_{t}, \hat{f}_{t}\right)+\operatorname{FFN}\left(\operatorname{\scriptstyle SWMSA}\left(\tilde{d}_{t}, \hat{f}_{t}\right)\right)\\
\hat{d}_{t}=\bar{d}_t+d_{t}
\end{array},
\label{eq:(3)}
\end{equation}
where $\tilde{d}_{t}$ is the output of the first-stage cross-attention of our STF module. For simplicity, the normalization operation is omitted in the above formulation.

\subsection{Temporal Modeling with GRViT Unit} \label{GRViT_temporal_modeling}

Temporal transformers have great success in various video sequence tasks~\cite{cao2022vdtr,li2021video}, which efficiently model temporal dependencies in a parallel manner. Nevertheless, one limitation is that these parallel processing temporal transformers require a large GPU memory. Another limitation is that the temporal information extracted in batch mode is limited. To overcome these limitations, we design a gate recurrent vision transformer (GRViT) unit that introduces a recursive mechanism into transformers, which can further improve temporal modeling capabilities for better performance while alleviating the expensive GPU memory cost.

The overview diagram of the proposed GRViT unit is shown in Fig.~\ref{fig:framework}\textcolor{red}{(b)}. For the current event temporal bin $S_{t}$, our GRViT unit  $\mathcal{G}$ takes the feature map $f_{t}$ and the memory hidden state $h_{t-1}$ from the previous temporal bin as the input, then outputs the current hidden state $h_{t}$ and the spatio-temporal feature map $\hat{f}_{t}$, and it can be formulated as:
\begin{eqnarray}
	(\hat{f}_{t}, h_{t}) = \mathcal{G}(f_{t}, h_{t-1}).
\end{eqnarray}

To be specific, our GRViT unit mainly consists of two core parts, namely the attention gate and the update gate. A learnable positional encoding vector needs to be appended to $f_{t}$ before inputting it into the GRViT unit. The attention gate is utilized to generate the attention feature map $A_{t}$. Firstly, $A_{t}$ is added to the input $f_{t}$ followed by a feed-forward network (FFN) with a residual connection, which is used to output the spatio-temporal feature map $\hat{f}_{t}$. Secondly, $A_{t}$ and $h_{t-1}$ are passed by the update gate and output the current memory hidden state $h_{t}$.

The attention gate aims at establishing the interaction between spatial and temporal domains from the current feature map and the previous hidden state. Firstly, the input of the attention gate is a triplet (i.e., $Q_{t}$, $K_{t}$, and $V_{t}$), which can be computed from $f_{t}$ and $h_{t-1}$ as:
\begin{equation}
\begin{aligned}
Q_{t}&=f_{t} W_{Q}^{f}+h_{t-1} W_{Q}^{h} \\
K_{t}&=f_{t} W_{K}^{f}+h_{t-1} W_{K}^{h} \\
V_{t}&=f_{t} W_{V}^{f}+h_{t-1} W_{V}^{h}
\end{aligned},
\label{eq:(4)}
\end{equation}
where $W_{Q}^{f}$, $W_{K}^{f}$, $W_{V}^{f}$, $W_{Q}^{h}$, $W_{K}^{h}$, and $W_{V}^{h}$ are learnable parameters of linear projection layers. Then, a linear attention operation replaces the SoftMax to prevent gradient vanishing, and it can be depicted as:
\begin{equation}
a_{t} = \left(elu\left(Q_{t}\right) + 1\right) \left(elu\left(K_{t}\right)^{\top} + 1\right) V_{t},
\label{eq:(5)}
\end{equation}
where $elu$ is the ELU activation function. Finally, the attention feature map $A_{t}$ can be obtained by an extension with $m$ independent linear-attention operations and project their concatenated outputs as:
\begin{equation}
A_{t} = \left[a_{t}^{1} ; ... ; a_{t}^{m}\right] W_{a},
\end{equation}
where $W_{a}$ denotes a linear layer that is used to project the attended vector.

As a result, the final output spatio-temporal feature map $\hat{f}_{t}$ can be formulated as:
\begin{equation}
\hat{f}_{t}=A_{t}+f_{t}+\operatorname{FFN}\left(A_{t}+f_{t}\right).
\label{eq:(6)}
\end{equation}

The update gate determines how much temporal clue will be passed to the next time step. $f_{t}$ and $h_{t-1}$ are concatenated and passed to a linear projection layer followed by a sigmoid function to output the gate $U_{t}$, which can be expressed as:
\begin{equation}
U_{t} = \sigma\left(\left[f_{t} ; h_{t-1}\right] W_{p}\right),
\label{eq:(7)}
\end{equation}
where $W_{p}$ refers to the linear projection layer, and $\sigma(\cdot)$ indicates the sigmoid activation function.

In fact, $U_{t}$ determines how much attended information to keep and how much temporal information in the previous hidden state to discard. Thus, the current hidden state $h_{t}$ can be computed as follows:
\begin{equation}
h_{t}=\left(1-U_{t}\right) \odot h_{t-1}+U_{t} \odot A_{t}.
\label{eq:(8)}
\end{equation}

\begin{table*}[t]
\small
\setlength{\tabcolsep}{0.81 mm}{
\begin{tabular}{llcccccccccc}
\toprule
Dataset        & \multicolumn{1}{c}{Method} & Abs.Rel. $\downarrow$ & RMSELog $\downarrow$ & SILog $\downarrow$ & $\delta<1.25 \uparrow$ & $\delta<1.25^{2} \uparrow$ & $\delta<1.25^{3} \uparrow$ & 10m $\downarrow$ & 20m $\downarrow$ & 30m $\downarrow$ & Runtime(ms) \\ \hline
\multirow{5}{*}{day1}                       & MDDE ~\cite{hidalgo2020learning}            & 0.450             & 0.514            & 0.251          & 0.472          & 0.711          & 0.823          & 2.70              & 3.46              & 3.84       & 7.67       \\
                                                     & DTL$-$ ~\cite{wang2021dual}            & 0.390             & 0.436            & 0.176          & 0.510          & 0.757          & 0.865          & 2.00              & 2.91              & 3.35       & 6.01       \\
                                                     & MDDE$+$ ~\cite{hidalgo2020learning}           & 0.346             & 0.421            & 0.172          & 0.567          & 0.772          & 0.876          & 1.85              & 2.64              & 3.13      & 7.67        \\
                                                     & DPT ~\cite{ranftl2021vision}             & {\ul 0.291}             & {\ul 0.341}            & {\ul 0.105}          & \textbf{0.668}    & {\ul 0.829}          & {\ul 0.914}          & {\ul 1.44}              & {\ul 2.40}              & {\ul 2.82}     & 24.51         \\  
                                                     & \textbf{EReFormer}          & \textbf{0.271}       & \textbf{0.333}      & \textbf{0.102}    & {\ul 0.664}          & \textbf{0.831}    & \textbf{0.923} & \textbf{1.29}        & \textbf{2.14}     & \textbf{2.59}    & 35.17    \\ \hline
\multirow{5}{*}{night1}                     & MDDE ~\cite{hidalgo2020learning}            & 0.770             & 0.638            & 0.346          & 0.327          & 0.582          & 0.732          & 5.36              & 5.32              & 5.40      & 7.67        \\
                                                     & DTL$-$ ~\cite{wang2021dual}            & 0.474             & 0.555            & 0.299          & 0.429          & 0.657          & 0.791          & 2.61             & 3.11              & 3.82       & 6.01       \\
                                                     & MDDE$+$ ~\cite{hidalgo2020learning}           & 0.591             & 0.646            & 0.374          & 0.408          & 0.615          & 0.754          & 3.38              & 3.82              & 4.46       & 7.67       \\
                                                     & DPT ~\cite{ranftl2021vision}             & {\ul 0.344}             & \textbf{0.405}   & \textbf{0.153} & \textbf{0.564} & \textbf{0.768} & \textbf{0.891} & {\ul 1.80}              & {\ul 2.67}              & {\ul 3.22}        & 24.51      \\ 
                                                     & \textbf{EReFormer}          & \textbf{0.317}       & {\ul 0.415}            & {\ul 0.158}          & {\ul 0.547}          & {\ul 0.753}          & {\ul 0.881}          & \textbf{1.52}        & \textbf{2.28}        & \textbf{2.98}  & 35.17  \\   \hline
\multirow{5}{*}{night2}                     & MDDE ~\cite{hidalgo2020learning}            & 0.400             & 0.448            & 0.176          & 0.411          & 0.720          & 0.866          & 2.80              & 3.28              & 3.74      & 7.67        \\
                                                     & DTL$-$ ~\cite{wang2021dual}            & 0.335             & 0.465            & 0.204          & 0.496          & 0.735          & 0.858          & 1.74              & {\ul 2.50}              & 3.29      & 6.01        \\
                                                     & MDDE$+$ ~\cite{hidalgo2020learning}           & 0.325             & 0.515            & 0.240          & 0.510          & 0.723          & 0.840          & {\ul 1.67}              & 2.63              & 3.58      & 7.67        \\
                                                     & DPT ~\cite{ranftl2021vision}             & {\ul 0.299}             & {\ul 0.362}            & {\ul 0.122}          & {\ul 0.610}          & {\ul 0.810}          & {\ul 0.915}          & 1.68              & 2.59              & {\ul 3.06}       & 24.51       \\  
                                                     & \textbf{EReFormer}          & \textbf{0.262}       & \textbf{0.346}      & \textbf{0.112}    & \textbf{0.619} & \textbf{0.826} & \textbf{0.927} & \textbf{1.40}        & \textbf{2.12}        & \textbf{2.66}  & 35.17   \\ \hline
\multicolumn{1}{c}{\multirow{5}{*}{night3}} & MDDE ~\cite{hidalgo2020learning}            & 0.343             & 0.410            & 0.157          & 0.451          & 0.753          & 0.890          & 2.39              & 2.88              & 3.39       & 7.67       \\
\multicolumn{1}{c}{}                                 & DTL$-$ ~\cite{wang2021dual}            & 0.307             & 0.458            & 0.195          & 0.501          & 0.734          & 0.860          & 1.54              & 2.37              & 3.26       & 6.01       \\
\multicolumn{1}{c}{}                                 & MDDE$+$ ~\cite{hidalgo2020learning}           & 0.277             & 0.424            & 0.162          & 0.541          & 0.761          & 0.890          & {\ul 1.42}              & {\ul 2.33}              & 3.18      & 7.67        \\
\multicolumn{1}{c}{}                                 & DPT ~\cite{ranftl2021vision}             & {\ul 0.272}             & {\ul 0.348}            & {\ul 0.116}          & \textbf{0.608} & {\ul 0.814}          & {\ul 0.920}          & 1.57              & 2.45              & {\ul 2.94}     & 24.51         \\  
\multicolumn{1}{c}{}                                 & \textbf{EReFormer}          & \textbf{0.248}       & \textbf{0.345}      & \textbf{0.109}    & {\ul 0.597}          & \textbf{0.818}    & \textbf{0.928}    & \textbf{1.32}        & \textbf{2.04}        & \textbf{2.68}   & 35.17     \\ \bottomrule
\end{tabular}}
\vspace{-0.2cm}
\caption{Quantitative results on the MVSEC dataset. These methods are all trained on the training split (i.e., outdoor day2), except for MDDE$+$ trained on both MVSEC and DENSE. $\downarrow$ indicates lower is better and $\uparrow$ higher is better. Our results outperform state-of-the-art event-based monocular depth estimators~\cite{hidalgo2020learning,wang2021dual} while obtaining better performance than the transformer-based frame-based method~\cite{ranftl2021vision}.}
\label{table:1}
 \vspace{-0.2cm}
\end{table*}

\begin{figure*}[t]
  \centering
  \includegraphics[width=\textwidth]{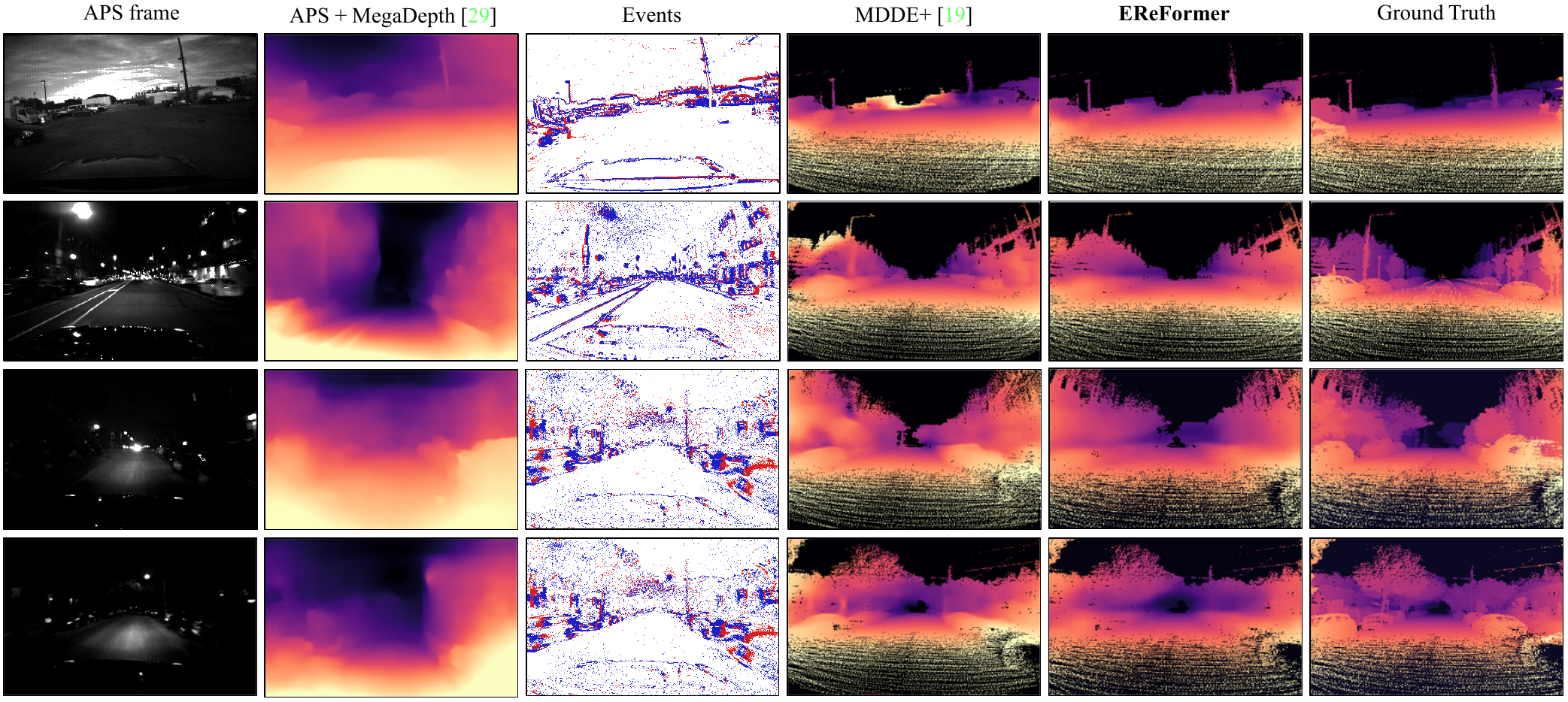}
  \caption{Representative examples of four test sequences in the MVSEC dataset. The first row to the fourth row corresponds to the outdoor day1, and outdoor night1 to outdoor night3, respectively. The second column refers to the MegaDepth~\cite{li2018megadepth} prediction using the APS frames. Note that, MegaDepth fails to predict the fine-grained depth map at low-light conditions. Compared with MDDE$+$~\cite{hidalgo2020learning}, our EReFormer can achieve more globally coherent predictions both day and night, which is closer to the ground truth.}
  \label{fig:MVSEC}
  \vspace{-0.2cm}
\end{figure*}

\begin{table*}[t]
\small
\setlength{\tabcolsep}{0.87 mm}{
\begin{tabular}{llcccccccccc}
\toprule
Dataset        & \multicolumn{1}{c}{Method} & Abs.Rel. $\downarrow$ & RMSELog $\downarrow$ & SILog $\downarrow$ & $\delta<1.25 \uparrow$ & $\delta<1.25^{2} \uparrow$ & $\delta<1.25^{3} \uparrow$ & 10m $\downarrow$ & 20m $\downarrow$ & 30m $\downarrow$ & Runtime(ms) \\ \hline
\multirow{3}{*}{Town06} & MDDE ~\cite{hidalgo2020learning}            & 0.120             & 0.188            & 0.035          & 0.855          & 0.956          & 0.987          & {\ul 0.31}        & 0.74              & 1.32      & 7.67        \\
                        & DTL$-$ ~\cite{wang2021dual}            & 0.211             & 0.280            & 0.078          & 0.706          & 0.897          & 0.963          & 0.95              & 1.40              & 1.98       & 6.01       \\
                        & DPT ~\cite{ranftl2021vision}             & {\ul 0.108}       & {\ul 0.170}      & {\ul 0.029}    & {\ul 0.881}    & \textbf{0.967} & \textbf{0.989} & 0.32              & \textbf{0.57}     & \textbf{1.08}   & 24.51   \\  
                        & \textbf{EReFormer}          & \textbf{0.095}    & \textbf{0.170}   & \textbf{0.029} & \textbf{0.881} & {\ul 0.960}    & {\ul 0.988}    & \textbf{0.21}     & {\ul 0.62}        & {\ul 1.15}    & 35.17    \\ \hline
\multirow{3}{*}{Town07} & MDDE ~\cite{hidalgo2020learning}            & 0.267             & 0.328            & 0.098          & 0.774          & 0.878          & 0.927          & 1.03              & 2.35              & 3.06       & 7.67       \\
                        & DTL$-$ ~\cite{wang2021dual}            & 0.334             & 0.375            & 0.111          & 0.625          & 0.809          & 0.895          & 1.44              & 2.95              & 3.60       & 6.01       \\
                        & DPT ~\cite{ranftl2021vision}             & {\ul 0.210}       & {\ul 0.294}      & {\ul 0.079}    & {\ul 0.778}    & {\ul 0.882}    & {\ul 0.938}    & {\ul 0.72}        & {\ul 1.73}        & {\ul 2.34}    & 24.51    \\  
                        & \textbf{EReFormer}          & \textbf{0.181}    & \textbf{0.276}   & \textbf{0.075} & \textbf{0.794} & \textbf{0.885} & \textbf{0.938} & \textbf{0.57}     & \textbf{1.50}     & \textbf{2.08}  & 35.17   \\ \hline
\multirow{3}{*}{Town10} & MDDE ~\cite{hidalgo2020learning}            & 0.220             & \textbf{0.323}   & \textbf{0.093} & 0.724          & \textbf{0.865} & \textbf{0.932} & 0.61              & 1.45              & 2.42      & 7.67        \\
                        & DTL$-$ ~\cite{wang2021dual}            & 0.259             & 0.416            & 0.162          & 0.588          & 0.776          & 0.869          & 0.84              & 1.46              & 2.16       & 6.01       \\
                        & DPT ~\cite{ranftl2021vision}             & {\ul 0.205}       & 0.356            & 0.101          & {\ul 0.733}    & 0.833          & 0.901          & {\ul 0.53}        & {\ul 1.04}        & {\ul 1.75}    & 24.51    \\  
                        & \textbf{EReFormer}          & \textbf{0.172}    & {\ul 0.335}      & {\ul 0.098}    & \textbf{0.747} & {\ul 0.839}    & {\ul 0.908}    & \textbf{0.29}     & \textbf{0.86}     & \textbf{1.49}  & 35.17   \\ \bottomrule
\end{tabular}}
\vspace{-0.20cm}
\caption{Quantitative results on the DENSE dataset. All methods are trained on the training split of DENSE. The first two sequences are used for validation and the Town10 sequence for testing. Our method outperforms state-of-the-art methods by a large margin in terms of average absolute depth errors metric on the testing set meanwhile achieves the minimal absolute relative error (Abs.Rel.).}
\label{table:2}
\vspace{-0.20cm}
\end{table*}

\begin{figure*}[t]
  \centering
  \includegraphics[width=1.00\linewidth]{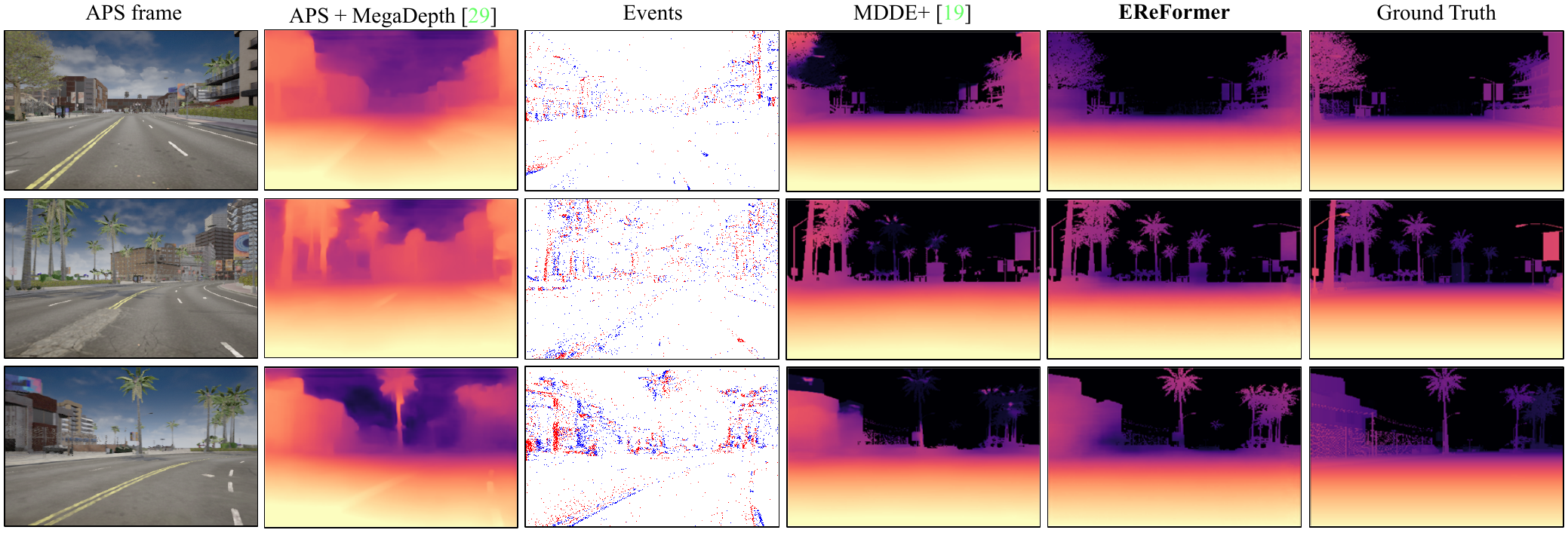}
  \caption{Representative examples of the testing sequence in the DENSE dataset. Obviously, our EReFormer obtains finer-grained and more globally coherent dense depth maps than the best event-based competitor that utilizes MDDE$+$~\cite{hidalgo2020learning} to process the event stream.}
  \label{fig:dense}
\end{figure*}

\section{Experiments}



\noindent \textbf{Datasets.} We report experimental results on a synthetic dataset (i.e., DENSE ~\cite{hidalgo2020learning}) and a real-world dataset (i.e., MVSEC~\cite{zhu2018multivehicle}). Following the previous work~\cite{hidalgo2020learning, wang2021dual,zhu2019unsupervised}, the DENSE dataset contains three subsets including Town01 to Town05 for training, Town06 and Town07 for validation, and Town10 for testing. For the MVSEC dataset, we use outdoor day2 for training, and four sequences (i.e., outdoor day1, and outdoor night1 to outdoor night3) for testing.

\noindent \textbf{Evaluation Metrics.} To compare different methods, absolute relative error (Abs.Rel.), logarithmic mean squared error (RMSELog), scale invariant logarithmic error (SILog), accuracy ($\delta<1.25^{n}, n=1,2,3$), average absolute depth errors at different cut-off depth distances (i.e., 10m, 20m and 30m), and running time ($ms$) are selected as five typical evaluation metrics, which are the most broadly utilized in the depth estimation task.

\noindent \textbf{Implementation Details.} Our EReFormer is implemented using the Pytorch framework~\cite{paszke2017automatic}. We use Swin-T ~\cite{liu2021swin} pre-trained on ImageNet as the backbone to achieve an accuracy-speed trade-off. We set the channel number $C$ to 96. During training, we use AdamW optimizer~\cite{kingma2014adam} with weight-decay 0.1 and set the 1-cycle policy~\cite{smith2019super} for the learning rate with $\text{max}\_\text{lr}=3.2\times10^{-5}$. We train our network for 200 epochs with batch size 2. All experiments are conducted on NVIDIA Tesla V100-PCLE GPUs.

\noindent \textbf{Comparisons.} To verify the effectiveness of the proposed approach, we compare EReFormer with four state-of-the-art methods (i.e.,  MDDE~\cite{hidalgo2020learning} for voxel grid, DTL$-$~\cite{wang2021dual} for event image, MDDE$+$~\cite{hidalgo2020learning} for voxel grid, and DPT~\cite{ranftl2021vision} for event image). It should be noted that MDDE$+$ is pretrained on the first 1000 samples in the DENSE dataset and then retrained on both two datasets, which shares the same architecture with MDDE.  DTL$-$ selects one branch of standard DTL~\cite{wang2021dual} to convert each event image into a depth map. DPT is an outstanding frame-based monocular depth estimator that adopts vision transformers to process each event image. To be fair, we evaluate DTL$-$ and the DPT architecture in the same experimental settings as ours. In addition, we will release the code upon acceptance.

\subsection{Effective Test}
\noindent \textbf{Evaluation on the MVSEC Dataset.} As is illustrated in Table~\ref{table:1}, we quantitatively compare our EReFormer with four state-of-the-art methods on the MVSEC dataset~\cite{zhu2018multivehicle}. All networks predict depth in the logarithmic scale, which is normalized and restored to absolute values by multiplying by the maximum depth clipped at 80 m. Note that, our EReFormer achieves the best performance across the whole test sets, especially the most valuable metric (i.e.,Abs.Rel.). At the same time, we can see that DPT~\cite{ranftl2021vision} using vision transformers obtains better performance than the best CNN-based method MDDE$+$~\cite{hidalgo2020learning}, which proves that utilizing the global spatial information from sparse events helps predict more accurate depth map in different scenarios. Although DPT has achieved satisfactory results for event-based monocular dense depth estimation, it is sub-optimal due to not leveraging rich temporal cues from continuous event streams. Compared the average absolute depth error of 10m, 20m, and 30m with DPT, our EReFormer achieves more accurate depth predictions at all distances with an average improvement overall test sequences of 14.8\% at 10m, 15.1\% at 20m, and 9.4\% at 30m with respect to values of DPT. In addition, our EReFormer is almost comparable to the computational speed of DPT. Overall, it can be concluded that efficient global sparse spatial modeling and temporal utilization can improve the performance of event-based monocular depth estimation. We further present some visualization results on the MVSEC Dataset in Fig.~\ref{fig:MVSEC}. Our EReFormer shows apparent advantages on the HDR scene when the APS frames (the second column) fail to predict the correct depth information in low-light conditions. Compared with the MDDE$+$, even if it was trained on both two datasets, our EReFormer trained only on the MVSEC dataset predicts finer-grained depth maps.

\noindent \textbf{Evaluation on the DENSE Dataset.} We further report quantitative results on the synthetic DENSE dataset~\cite{hidalgo2020learning} to validate the effectiveness of our EReFormer. As shown in Table~\ref{table:2}, our EReFormer achieves the best absolute relative error (Abs.Rel.) on three sequences. Meanwhile, our EReFormer improves the average absolute depth error about by 45.3\% at 10m, 17.3\% at 20m, and 14.9\% at 30m with respect to values of DPT on test sequence Town10. Besides, we find that our approach is sub-optimal in some metrics on validation sequence Town06. This is because the distribution of all scenarios in Town06 is too monolithic. From Fig.~\ref{fig:dense}, some visualization examples show that our EReFormer obtains higher-quality depth maps over the best event-based competitor (i.e., MDDE$+$~\cite{hidalgo2020learning}).


\subsection{Ablation Test}

Beyond effective tests, we next conduct ablation tests on the MVSEC dataset (e.g., outdoor day1 sequence) to take a deep look at the impact of each design choice.

\begin{table}[t]
\small
\setlength{\tabcolsep}{2.9 mm}{
\begin{tabular}{lcccc}
\toprule
Method      & The baseline & (a)   & (b)   & Ours   \\ \hline
STF         &              & $\boldsymbol{\checkmark}$     &       & $\boldsymbol{\checkmark}$              \\
GRViT unit &              &       & $\boldsymbol{\checkmark}$     & $\boldsymbol{\checkmark}$              \\ \hline
Abs. Rel. $\downarrow$  & 0.305        & 0.295 & {\ul 0.280} & \textbf{0.271} \\
Runtime(ms)   & 27.43        & 33.68 & 29.03 & 35.17 \\ \bottomrule
\end{tabular}}
\vspace{-0.20cm}
\caption{Performance components of our EReFormer on the outdoor day1 sequence. The baseline implement a transformer-based encoder-decoder network with a ADD skip connection.}
\label{table:3}
\end{table}

\begin{table}[t]
\small
\setlength{\tabcolsep}{0.7 mm}{
\begin{tabular}{lccccc}
\toprule
Fusion methods    & 10m $\downarrow$           & 20m $\downarrow$           & 30m $\downarrow$           & Abs.Rel. $\downarrow$    & Runtime(ms)   \\ \hline
ADD            & {\ul 1.47}    & {\ul 2.55}    & {\ul 3.02}    & {\ul 0.305}  & 27.43  \\
CONCAT & 1.49          & 2.58          & 3.07          & 0.307      & 32.96    \\
STF         & \textbf{1.41} & \textbf{2.42} & \textbf{2.91} & \textbf{0.295}  & 33.68 \\ 
\bottomrule
\end{tabular}}
\vspace{-0.20cm}
\caption{Comparison with typical skip connection strategies including the operations of ADD and CONCAT.}
\label{table:4}
\end{table}

\begin{table}[t]
\small
\setlength{\tabcolsep}{0.26 mm}{
\begin{tabular}{lccccc}
\toprule
Transfer methods    & 10m $\downarrow$           & 20m $\downarrow$           & 30m $\downarrow$           & Abs.Rel. $\downarrow$   & Runtime(ms)   \\ \hline
Attended            & {\ul 1.41}    & {\ul 2.28}    & {\ul 2.73}    & {\ul 0.290}   & 35.14 \\
Residual connection & 1.69          & 2.58          & 2.91          & 0.330         & 35.15 \\
Update gate         & \textbf{1.29} & \textbf{2.14} & \textbf{2.59} & \textbf{0.271} & 35.17 \\ 
\bottomrule
\end{tabular}}
\vspace{-0.20cm}
\caption{Depth estimation performance with different hidden state transfer operations in the GRViT unit.}
\label{table:5}
\vspace{-0.30cm}
\end{table}

\noindent \textbf{Contribution of Each Component.} As shown in Table~\ref{table:3}, two methods, namely (a) and (b), utilize a spatial transformer fusion (STF) module to fuse multi-scale features, and a gate recurrent vision transformer (GRViT) unit for temporal modeling, consistently achieve better performance on the outdoor day1 sequence than the baseline using the transformer-based encoder-decoder backbone. More precisely, compared (a) and the baseline, the absolute promotion is 3.3\%, which demonstrates that it is feasible to adopt the STF module between the encoder and decoder sub-networks. Our GRViT unit, leveraging temporal cues, obtains the 8.2\% Abs.Rel. improvement over the baseline. Besides, the last row of Table~\ref{table:3} shows that the computational speeds of these methods are almost comparable.

\noindent \textbf{Comparison with Skip Connection Strategies.} We compare the STF module in transformer-based encoder-decoder networks with some typical operations (e.g., ADD and CONCAT) in Table~\ref{table:4}. Notably, our STF module achieves the best performance against the ADD and CONCAT operations while keeping almost comparable computational costs. For example, our strategy obtains finer-grain predictions at all distances with an improvement of $4.1\%$ at 10m, $5.1\%$ at 20m, and $3.6\%$ at 30m with the ADD operation.

\noindent \textbf{Ablating Hidden State Transfer.} We compare the update gate in our GRViT unit with two typical methods in Table~\ref{table:5}. The attended operation only uses the attended feature $A_{t}$ from the attention gate as the current hidden state $h_{t}$. The residual operation adds a residual connection between the hidden state $h_{t-1}$ and the attended feature $A_{t}$. From Table~\ref{table:5}, we find that the residual connection achieves worse results because the temporal information from a long time window is not forgotten. On the contrary, our update gate outperforms two compared transfer methods while maintaining almost comparable computational speed.

\section{Discussion}
\begin{figure}[t]
  \centering
  \includegraphics[width=1.00\linewidth]{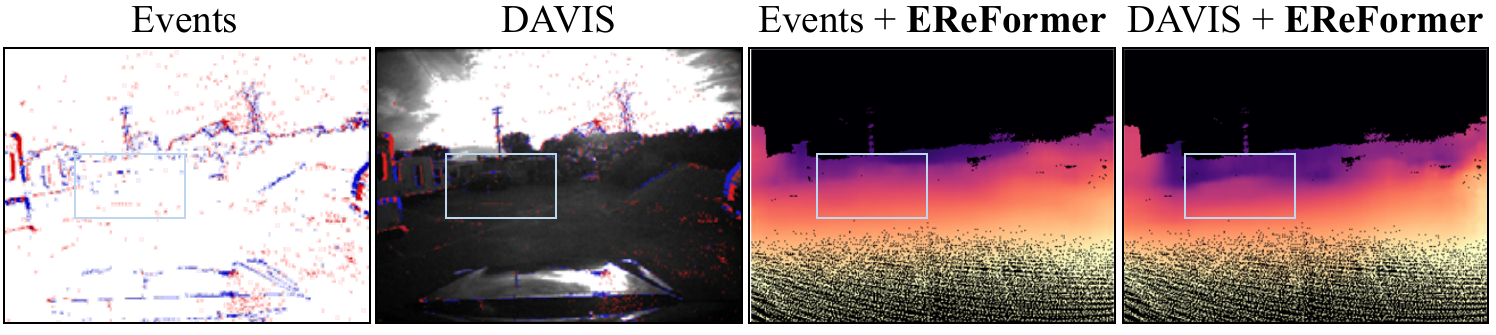}
  \caption{A failure case of our EReFormer in the slow-moving scenario. Our EReFormer is hard to generate dense depth maps without enough events. Notably, the usage of auxiliary frames can improve the performance of monocular depth estimation.}
  \label{fig:discussion}
  \vspace{-0.4cm}
\end{figure}

\noindent \textbf{Limitation.} Although our EReFormer achieves satisfactory results even in challenging scenes, some failure cases still remain. As depicted in Fig.~\ref{fig:discussion}, the first and third columns show that the slow-moving scene is hard to perform high-quality depth prediction. This is because event cameras evidently sense dynamic changes, but they generate almost no events in static or slow-moving scenarios.

\noindent \textbf{Opportunity.} The last two columns in Fig.~\ref{fig:discussion} indicate that the DAVIS camera~\cite{brandli2014240}, streaming two complementary modalities of events and frames, providing a viable solution to address the above limitation. In fact, how to design a pure transformer to integrate events and frames for dense depth estimation is a worthwhile topic in the future.

\section{Conclusion}
This paper presents a novel event-based monocular depth estimator with recurrent transformers (i.e., EReFormer), which effectively models global sparse spatial context information and leverages rich temporal cues from a continuous event stream. To the best of our knowledge, this is the first work to explore such a pure transformer to predict dense depth maps for a monocular event camera. Our EReFormer consists of two core modules, namely a spatial fusion transformer (STF) and a gate recurrent vision transformer (GRViT). The results show that our EReFormer outperforms state-of-the-art methods by a margin on both synthetic and real-world datasets. We believe that our EReFormer acts as a bridge between event cameras and practical applications involving monocular depth estimation, especially in fast-motion and low-light scenarios.


{\small
\bibliographystyle{ieee_fullname}
\bibliography{manuscript}
}

\end{document}